\documentclass[letterpaper]{article} 

\usepackage{uai2020}  
\usepackage[margin=1in]{geometry}
\usepackage{times}
\usepackage[sort&compress, numbers]{natbib}

\usepackage{amsthm,amssymb,amsmath}
\usepackage{mathtools}
\usepackage{breqn}
\usepackage{subfigure}

\usepackage[ruled, linesnumbered]{algorithm2e}

\newcommand{\overbar}[1]{\mkern 1.5mu\overline{\mkern-1.5mu#1\mkern-1.5mu}\mkern 1.5mu}

\newcommand{\E}{\mathbb{E}}
\newcommand{\A}{\mathcal{A}}
\newcommand{\Sc}{\mathcal{S}}
\newcommand{\As}{\mathcal{A}_s}

\newcommand{\wt}{\omega_{1:t}}
\newcommand{\wk}{\overbar{\omega}_{1:k}}
\newcommand{\Wk}{\Omega_{1:k}}

\newcommand{\wopt}{\overbar{\omega}^*}
\newcommand{\wbar}{\overbar{\omega}}

\newcommand{\VOC}{\textsc{VOC}}
\newcommand{\VOCp}{\textsc{VOC$'$}}

\newcommand{\nsr}{(s', a') \in \Gamma_n(s)}

\theoremstyle{lemma}

\theoremstyle{theorem}

\newenvironment{hproof}{%
  \proof}{\endproof}

\theoremstyle{proposition}
\newtheorem{proposition}{Proposition}

\theoremstyle{remark}

\theoremstyle{definition}
\newtheorem{definition}{Definition}

\theoremstyle{corollary}

\theoremstyle{definition}

\title{Static and Dynamic Values of Computation in MCTS}

\author{\textbf{Eren Sezener} \\ DeepMind \And \textbf{Peter Dayan}\\Max Planck Institute for Biological Cybernetics T{\"u}bingen\\
University of T{\"u}bingen}

\begin{document}

\maketitle

\begin{abstract}

Monte-Carlo Tree Search (MCTS) is one of the most-widely used methods
for planning, and has powered many recent advances in artificial
intelligence. In MCTS, one typically performs computations
(i.e., simulations) to collect statistics about the possible future
consequences of actions, and then chooses accordingly. Many
popular MCTS methods such as UCT and its variants decide which
computations to perform by trading-off exploration and exploitation. In
this work, we take a more direct approach, and explicitly quantify the
value of a computation based on its expected impact on the quality of
the action eventually chosen. Our approach goes beyond the \emph{myopic}
limitations of existing computation-value-based methods in two senses:
(I) we are able to account for the impact of non-immediate (ie, future)
computations (II) on non-immediate actions. We show that policies that
greedily optimize computation values are optimal under certain
assumptions and obtain results that are competitive with the
state-of-the-art.

\end{abstract}

\section{INTRODUCTION}

Monte Carlo tree search (MCTS) is a widely used approximate planning method
that has been successfully applied to many challenging domains such as computer
Go \citep{coulom2007,silver2016}. In MCTS, one estimates values of actions by
stochastically expanding a search tree---capturing potential future states and
actions with their respective values. Most MCTS methods rely on rules
concerning how to expand the search tree, typically trading-off exploration and
exploitation such as in UCT \citep{kocsis2006}. However, since no ``real'' reward accrues during internal search, UCT can be viewed as a heuristic \citep{hay2011,hay2012,tolpin2012}. In this paper, we propose a more
direct approach by calculating values of MCTS computations (i.e., tree
expansions/simulations).

In the same way that the value of an action in a Markov decision process (MDP)
depends on subsequent actions, the value of a computation in MCTS should
reflect subsequent computations. However, computing the optimal computation
values---the value of a computation under an optimal computation policy---is
known to be generally intractable \citep{lin2015,wefald1991}. 
Therefore, one often resorts to ``myopic'' approximations of computation values, such as
considering the impact of only the immediate computation in isolation from the subsequent computations. For instance, it has been shown that simple modifications to UCT, where myopic computation values inform the tree policy at the root node, can yield significant improvements \cite{hay2012,tolpin2012}.

In this work, we propose tractable yet non-myopic methods for calculating computation values,
going beyond the limitations of existing methods. To this end, we introduce static and dynamic value functions that form lower and upper bounds for state-action values in MCTS, independent of any future computations. We then utilize these functions to define static and dynamic values of computation, capturing the expected change in state values resulting from a computation. We show that the existing myopic computation value definitions in MCTS can be seen as specific instances of static computation values. The dynamic value function, on the other hand, is novel measure, and it enables non-myopic ways of selecting MCTS computations. We prove that policies that greedily maximize static/dynamic computation values are asymptotically optimal under certain assumptions. Furthermore, we also show that they outperform various MCTS baselines empirically.

\section{BACKGROUND}

In this section we cover some of relevant literature and introduce the notation.

\subsection{MONTE CARLO TREE SEARCH}
MCTS algorithms function by incrementally and stochastically building a search
tree (given an environment model) to approximate state-action values. This incremental growth prioritizes
the promising regions of the search space by directing the growth of the tree
towards high value states. To elaborate, a \emph{tree policy} is used to
traverse the search tree and select a node which is not fully
expanded---meaning, it has immediate successors that aren't included in the
tree. Then, the node is expanded once by adding one of its unexplored children
to the tree, from which a trajectory simulated for a fixed number of steps or
until a terminal state is reached. Such trajectories are generated using a
\emph{rollout policy}; which is typically fast to compute---for instance random
and uniform. The outcome of this trajectory---i.e., cumulative discounted
rewards along the trajectory---is used to update the value estimates of the
nodes in the tree that lie along the path from the root to the expanded node.

Upper Confidence Bounds applied to trees (UCT) \citep{kocsis2006} adapts a
multi-armed bandit algorithm called UCB1 \citep{auer2002} to MCTS. More
specifically, UCT's tree policy applies the UCB1 algorithm recursively down the
tree starting from the root node. At each level, UCT selects the most promising
action at state $s$ via $\arg \max_{a \in \mathcal{A}_s} \hat{Q}(s,a) + c
\sqrt{\frac{2 \log N(s)}{N(s,a)}}$ where $\mathcal{A}_s$ is the set of
available actions at $s$, $N(s,a)$ is the number of times the $(s,a)$ is
visited, $N(s) \coloneqq \sum_{a \in \mathcal{A}_s}N(s,a)$, $\hat{Q}(s,a)$ is
the average reward obtained by performing rollouts from $(s,a)$ or one of its
descendants, and $c$ is a positive constant, which is typically selected
empirically. The second term of the UCT-rule assigns higher scores to nodes
that are visited less frequently. As such, it can be thought of as an
exploration bonus.

UCT is simple and has successfully been utilized for many applications.
However, it has also been noted \cite{tolpin2012,hay2012} that UCT's goal is
different from that of approximate planning. UCT attempts to ensure that the
agent experiences little \emph{regret} associated with the actions that are
taken during the Monte Carlo simulations that comprise planning. However, since
these simulations do not involve taking actions in the environment, the agent
actually experience no true regret at all. Thus failing to explore actions
based on this consideration could slow down discovery of their superior or
inferior quality.

\subsection{METAREASONING \& VALUE OF INFORMATION}

\citet{howard1966} was the first to quantify mathematically the economic gain
from obtaining a piece of information. \citet{wefald1991} formulated the
\emph{rational metareasoning} framework, which is concerned with how one should
assign values to meta-level actions (i.e., computations).
\citet{hay2012,tolpin2012} applied the principles of this framework to MCTS 
by modifying the tree-policy of UCT at the root node such that the selected
child node maximizes the value of information. They showed empirically
that such a simple modification can yield significant improvements. 

The field of Bayesian optimization has evolved in parallel.
For instance, what are known as \emph{knowledge gradients}
\citep{ryzhov2012,frazier2017} are equivalent to information/computation value formulations for flat/stateless problems such as
multi-armed bandits.

Computation values have also been used to explain  human and animal
behavior. For example, it has been suggested that humans might leverage
computation values to solve planning tasks in a resource efficient manner
\citep{lieder2014,sezener2019}, and animals might improve their
policies by ``replaying'' memories with large computation values
\citep{mattar2018}.

\subsection{NOTATION}

A finite Markov decision process (MDP) is a $5$-tuple $( \Sc, \A, \mathcal{P},
\mathcal{R}, \gamma )$, where $\Sc$ is a finite set of states $\A$ is a finite
set of actions, $\mathcal{P}$ is the transition function such that
$\mathcal{P}^a_{ss'} = P(s_t = s' | s_t = s, a_t = a)$, where $s,s' \in \Sc$
and $a \in \A$, $\mathcal{R}$ is the expected immediate reward function such
that $\mathcal{R}^a_{ss'} = \E[r_{t+1} | s_t = s, a_t = a, s_{t+1} = s']$,
where again $s,s' \in \Sc$ and $a \in \A$, $\gamma$ is the discount factor such
that $\gamma \in [0, 1)$.

We assume an agent interacts with the environment via a (potentially
stochastic) policy $\pi$, such that $\pi(s,a) = P(a_t = a | s_t = s)$. These
probabilities typically depend on parameters; these are omitted from the
notation. The value of an action $a$ at state $s$ is defined as the expected
cumulative discounted rewards following policy $\pi$, that is $Q^\pi(s, a) =
\mathbb{E}_\pi \left[ \sum_{i=0}^{\infty}{\gamma}^i r_{t+i} \right | s_t = s,
a_t=a]$.

The optimal action value function is defined as $Q^*(s, a) = \max_\pi
Q^\pi(s,a)$ for all state-action pairs, and satisfies the
\emph{Bellman optimality recursion}:
\begin{equation*}
Q^*(s,a) = \sum_{s'}\mathcal{P}_{ss'}^{a} \left[ \mathcal{R}_{ss'}^{a} +
\gamma \max_{a'} Q^*(s', a') \right] \: . 
\end{equation*}
We use $\mathcal{N}(\mu, \Sigma)$ and $\mathcal{N}(\mu, \sigma^2)$ to denote a
multivariate and univariate Normal distribution respectively with mean
vector/value $\mu$ and covariance matrix $\Sigma$ or scale $\sigma$.

\section{STATE-ACTION VALUES IN MCTS}

\begin{figure}[h]
\begin{center}
\includegraphics[width=0.35\textwidth]{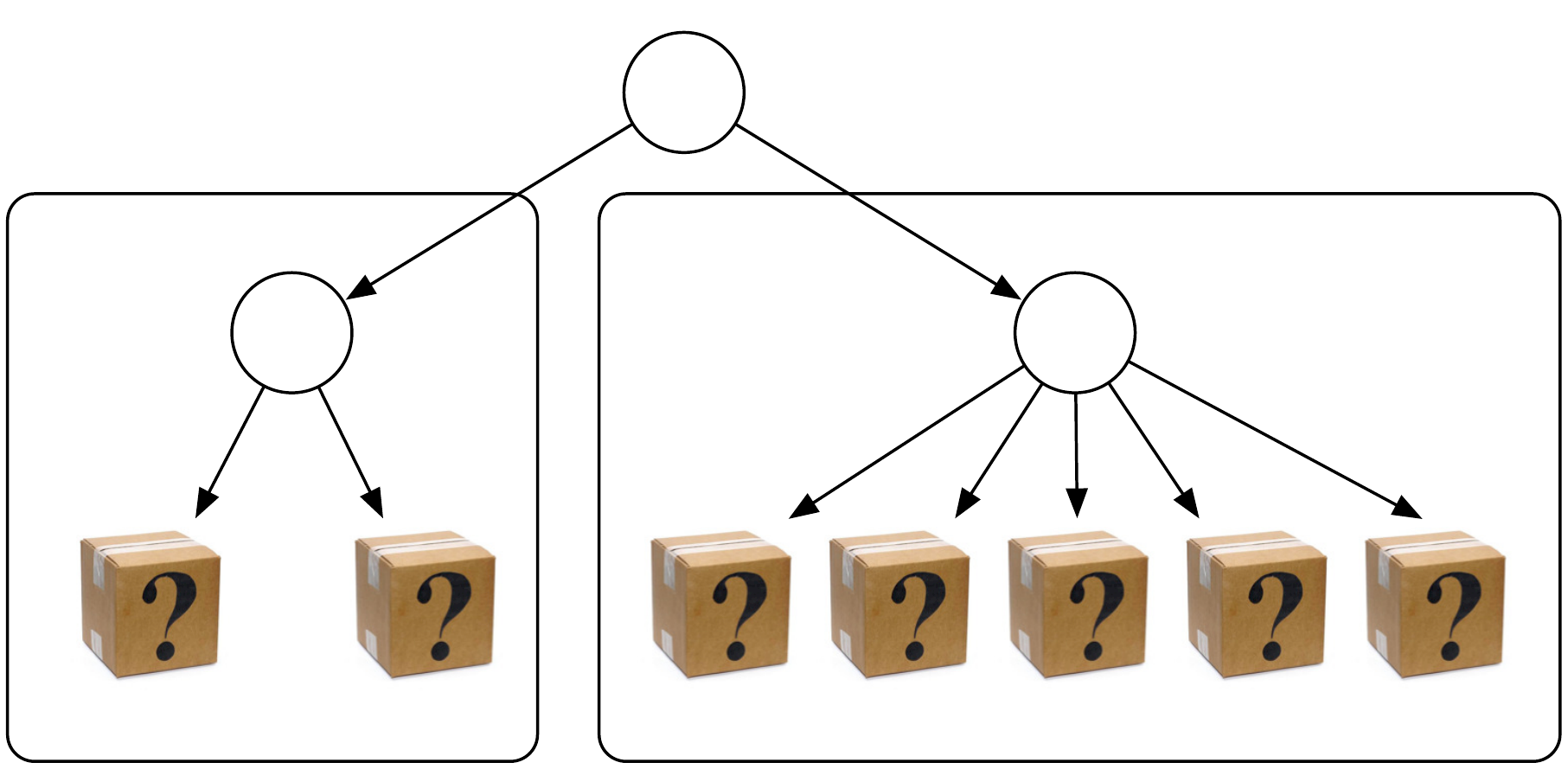}
\caption{Illustration of the example. There are two rooms, one containing two boxes, another one containing five boxes. There is an unknown amount of money in each box.}
\label{optionality}
\end{center}
\end{figure}

To motivate the issues underlying this paper, consider the following example
(Figure~\ref{optionality}). Here, there are two rooms: one containing two boxes
and the other containing five boxes. Each box contains an unknown but i.i.d.
amount of money; and you are ultimately allowed to open only one box. However,
you do so in stages. First you must choose a room, then you can open one of the
boxes and collect the money. Which room should you choose? What if you know
ahead of time that you could peek inside the boxes \emph{after} choosing the
room?

In the first case, it doesn't matter which room one chooses, as all the
boxes are equally valuable in expectation in absence of any further
information.  By contrast, in the second case, choosing the room
with five boxes is the better option. This is because one can obtain
further information by peeking inside the boxes---and more boxes mean
more money in expectation, as one has the option to choose the best one.

Formally, let $X = \{x_i\}_{i=1}^{n_x}$ and $Y = \{y_i\}_{i=1}^{n_y}$ be
sets of random variables denoting rewards in the boxes of the first and
the second room respectively. Assume all the rewards are sampled
i.i.d. and $n_x < n_y$.  Then we have $\max_{x \in X} \E[x]
= \max_{y \in Y} \E[y]$, which is why the two rooms are equally valuable
if one has to choose a box blindly. On the other hand, $\E[\max_{x \in
X} x] < \E[ \max_{y \in Y} y]$, which is analogous to the case where
boxes can be peeked in first.

If we consider MCTS with this example in mind, when we want to value an action
at the root of the tree, backing up the estimated mean values of the actions
lower in the tree may be insufficient. This is because the value of a root
action is a convex combination of the ``downstream'' (e.g., leaf) actions; and,
as such, uncertainty in the values of the leaves contributes to the expected
value at the root due to Jensen's inequality. We formalize this notion of value
as \emph{dynamic value} in the following section and utilize it to define
computation values later on.

\subsection{STATIC AND DYNAMIC VALUES} 

We assume a planning setting where the environment dynamic (i.e., $\mathcal{P}$
and $\mathcal{R}$) is known. We could then compute $Q^*$ in principle; however,
this is typically computationally intractable. Therefore, we estimate $Q^*$ by
performing computations such as random environment simulations (e.g., MCTS
rollouts). Note that, our uncertainty about $Q^*$ is not
epistemic---environment dynamic is known---but it is computational. In other
words, if we do not know $Q^*$, it is because we haven't performed the
necessary computations. In this subsection, we introduce static and dynamic
value functions, which are ``posterior'' estimates of $Q^*$ conditioned on
computations.

Let us unroll the Bellman optimality equation for $n$-steps\footnote{Obtaining,
what is sometimes referred to as the $n$-step Bellman equation.}. For a given
``root'' state, $s_{\rho}$, let $\Gamma_{n}(s_{\rho})$ be the set of leaf
state-actions---that is, state-actions can be transitioned to from $s_{\rho}$
in exactly $n$-steps. Let $Q_0^*$ be a random function denoting our prior beliefs about the optimal value function $Q^*$
over $\Gamma_{n}(s_{\rho})$. We then use $Q_n^*(s,
a)$ to denote the (Bayes-)optimal state-action value, which we define
as a function of $Q_0^*$:
\begin{align*}
Q_n^*(s, a) = \begin{cases}
Q_0^*(s,a) & \text{if $n = 0$}\\
\sum_{s'} \mathcal{P}_{ss'}^a [ \mathcal{R}_{ss'}^{a} + &\text{else} \\
\quad  \gamma \max_{a' \in \A_{s'}} Q^*_{n-1}(s', a') ]\\
\end{cases} \: ,
\end{align*}
where $\A_{s'}$ is the set of actions available at $s'$.

We assume it is possible to obtain noisy evaluations of $Q^*$ for leaf
state-actions by performing computations such as trajectory simulations. We
further assume that the process by which a state-action value is sampled is a
given, and we are interested in determining which state-action to sample from.
Therefore, we associate each computation with a single state-action in
$\Gamma_{n}(s_{\rho})$; but, the outcome of a computation might be informative
for multiple leaf values if they are dependent. Let $\wbar \coloneqq (s,a) \in
\Gamma_{n}(s_{\rho})$ be a \emph{candidate computation}. We denote the unknown
outcome of this computation at time $t$ with random variable $O_{\wbar t}$ (or
equivalently, $O_{sat}$), which we assume to be $O_{\wbar t} = Q_0^*(\wbar) +
\epsilon_t$ where $\epsilon_t$ is an unknown noise term with a known
distribution and is i.i.d. sampled for each $t$. If we associate a candidate
computation $\wbar$ with its unknown outcome $O_{\wbar t}$ at time $t$, we
refer to the resulting tuple as a \emph{closure} and denote it as $\Omega_{t} \coloneqq
(\wbar, O_{\wbar t})$. Finally, we denote a \emph{performed computation} at
time $t$, by dropping the bar, as $\omega_{t} \coloneqq (\wbar, o_{\wbar t})$
where $o_{\wbar t}$ (or equivalently, $o_{sat}$) is the observed outcome of the
computation that we assume to be $o_{\wbar t} \sim O_{\wbar t}$, and thus
$\omega_{t} \sim \Omega_{t}$. In the context of MCTS, $o_{sat}$ will be the
cumulative discounted reward of a simulated trajectory from $(s, a)$ at time
$t$. We will obtain these trajectories using an adaptive, asymptotically
optimal sampler/simulator (e.g., UCT), such that $\lim_{t \rightarrow \infty}
\E[O_{sat}] = Q^*(s,a)$. This means $\{O_{sat}\}_t$ is a non-stationary
stochastic process in practice; yet, we will treat it as a stationary process,
as reflected in our i.i.d. assumption.

Let $\wt$ be a sequence of $t$ performed computations concerning arbitrary
state-actions in $\Gamma_{n}(s_{\rho})$ and $s_{\rho}$ be the current state of
the agent on which we can condition $Q_0^*$. Because $\wt$ contains the
necessary statistics to compute the posterior leaf values, we will sometimes refer to it as
the \emph{knowledge state}. We denote the resulting posterior values for a $(s,a) \in \Gamma_{n}(s_{\rho})$ as $Q_0^*(s, a) | \wt$ and the joint values of leaves as $Q_0^* | \wt = (Q_0^*(s,a)
| \wt : ( s,a ) \in \Gamma_{n}(s_{\rho}))$.

We define the \emph{dynamic value function} as the expected value the agent
should assign to an action at $s_{\rho}$ given $\wt$, assuming it could resolve
all of the remaining uncertainty about posterior leaf state-action values
$Q_0^* | \wt$.

\begin{definition} The \textbf{dynamic value function} is defined as $\psi_n(s, a | \wt) \coloneqq
\mathbb{E}_{Q_0^* | \wt}\left[ \Upsilon_n(s,a | \wt) \right]$ where,
\begin{equation*}
\Upsilon_n (s, a | \wt) \coloneqq
\begin{cases}
Q_0^*(s,a) | \wt&  \text{if $n = 0$}\\
\sum_{s'} \mathcal{P}_{ss'}^a[ \mathcal{R}_{ss'}^{a} + & \text{else}\\
\; \gamma \max_{a'} \Upsilon_{n-1}(s', a' | \wt) ]\\
\end{cases}
\end{equation*}
where $\mathcal{A}_{s'}$ is the set of actions available at $s'$.
\end{definition}

The `dynamic' in the term reflects the fact that the agent may change its mind
about the best actions available at each state within $n$-steps; yet, this is
reflected and accounted for in $\psi_n$.
A useful property of $\psi_n$ is that is time-consistent in the sense that it
does not change with further computations in expectation. Let $\Wk \coloneqq
\{(\wbar_i, O_{\wbar_i i})\}_{i=1}^{k}$ be a sequence of $k$ closures.
Then the following equality holds for any $\Wk$:
\begin{equation}
\psi_n({s_{\rho}},a | \wt) = \E_{\Wk}[\psi_n({s_{\rho}},a | \wt  \Wk)] \: ,
\label{consistency}
\end{equation}
due to the law of total expectation, where $\wt \Wk$ is a concatenation. This
might seem paradoxical: why perform computations if action values do not change
in expectation? The reason is that we care about the maximum of dynamic values
over actions at ${s_{\rho}}$, which increases in expectation as long as
computations resolve some uncertainty. Formally, $ \max_{a \in
\mathcal{A}_{s_{\rho}}} \psi_n({s_{\rho}},a | \wt) \le \E_{\Wk}[\max_{a \in
\mathcal{A}_{s_{\rho}}}\psi_n({s_{\rho}},a | \wt \Wk)]$, due to Jensen's
inequality, just as in the example of the boxes.

Dynamic values capture one extreme: valuation of actions assuming perfect
information in the future. Next, we consider the other extreme, valuation under
zero information in the future, which is given by the \emph{static value
function}.

\begin{definition} We define the \textbf{static value function} as
\begin{equation*}
\phi_n (s, a | \wt) \coloneqq \begin{cases}
\mathbb{E}\left[ Q_0^*(s,a) | \wt \right]& \text{if $n = 0$}\\
\sum_{s'} \mathcal{P}_{ss'}^a [ \mathcal{R}_{ss'}^{a} +& \text{else}\\
\;  \gamma \max_{a'} \phi_{n-1}(s', a' | \wt) ]\\
\end{cases}
\end{equation*}
where $\mathcal{A}_{s'}$ is the set of actions available at $s'$.
\end{definition}

\begin{figure*}[t]
\centering
\includegraphics[width=1\textwidth]{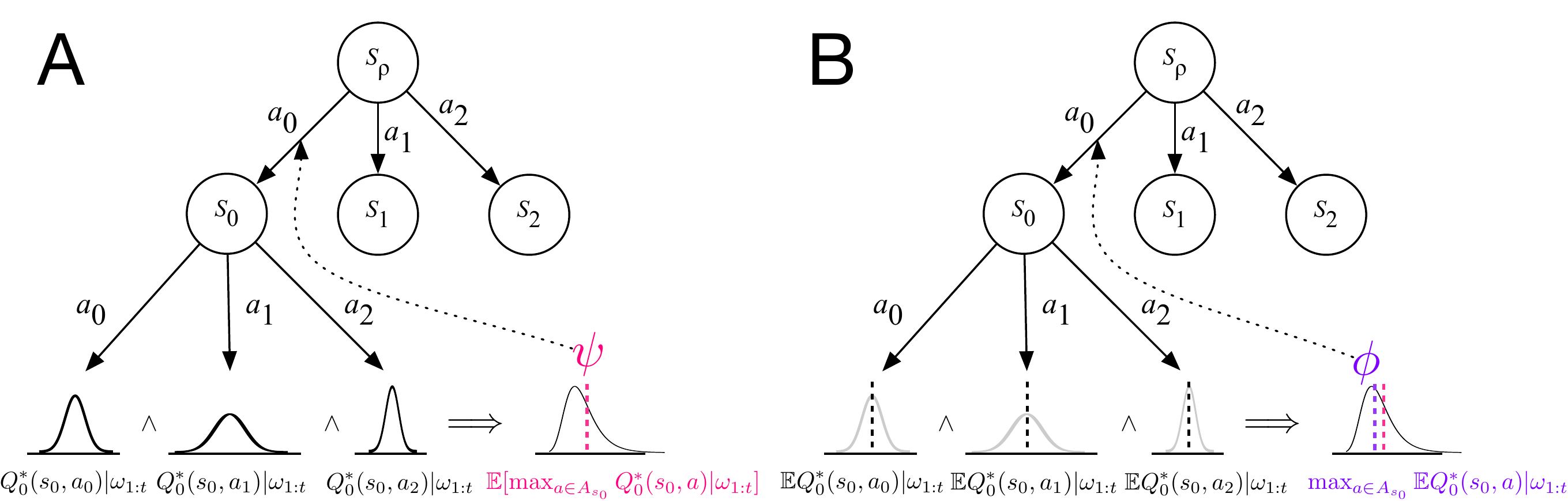}
\caption{Graphical illustration dynamic ($\psi_n$) and static ($\phi_n$) value functions for $n=2$. We ignore immediate rewards and the discounting for simplicity. In Panel~A, dynamic values (given by $\psi_2$) are obtained by calculating the \emph{expected maximum} of all state-action values (given by $Q_0^*$) lying $2$-steps away. Whereas, the static values (given by $\phi_2$) are obtained by calculating the \emph{maximum of expectations} of state-action values, as shown in Panel~B.}
\label{phi_psi}
\end{figure*}

In other words, $\phi_n({s_{\rho}}, a)$ captures how valuable $({s_{\rho}},a
)$ would be if the agent were to take $n$ actions before running any new
computations. In Figure~\ref{phi_psi}, we graphically contrast dynamic and
static values, where the difference is the stage at which the expectation is
taken. For the former, it is done at the level of the root actions; for the
latter, at the level of the leaves. 

Going back to our example with the boxes,
dynamic value of a room assumes that you open all the boxes after entering the
room, whereas the static value assumes you do not open any boxes. 
What can we
say about the in-between cases: action values under a finite number of future computations?
Assume we know that the agent will perform $k$ computations before taking an
action at $s_{\rho}$. The optimal allocation of these $k$ computations to leaf
nodes is known to be intractable even in a simpler bandit setting
\citep{madani2004}. That said, for any allocation (and for any finite $k$),
static and dynamic values will form lower and upper bounds on expected action
values nevertheless. We formalize this for a special case below. 

\begin{proposition}
Assume an agent at state $s_{\rho}$ and knowledge state $\wt$ decides to perform $\wk$, a sequence of $k$ candidate computations,
before taking $n$ actions. Then the expected future value of $a \in \mathcal{A}_{s_\rho}$ prior to observing any of
the $k$-computation outcomes is equal to $\E_{\Wk}[\phi_n({s_{\rho}}, a | \wt
\Wk)]$, where $\Wk = \{(\wbar_i, O_{\wbar_i i})\}_{i=1}^{k}$. Then,
\begin{equation*}
\psi_n(s_{\rho}, a | \wt) \ge \E_{\Wk}[\phi_n({s_{\rho}}, a | \wt \Wk)] \ge \phi_n(s_{\rho}, a | \wt) \: ,
\end{equation*}
where both bounds are tight.
\label{prop:bounds}
\end{proposition}
The proof is provided in the Appendix.

\section{VALUE OF COMPUTATION} 

In this section, we use the static and dynamic value functions to define computation values.
We show that these computation values have desirable properties and that policies greedily-maximizing these values are optimal in certain senses. 
Lastly, we compare our definitions to the existing computation value definitions.

\begin{definition} We
define the \textbf{value of computation} at state ${s_{\rho}}$ for a sequence
of candidate computations $\wk$ given a static or dynamic value function $f \in
\{\phi_n, \psi_n\}$ and a knowledge state $\wt$ as 
\begin{align*}
\VOC_{f}({s_{\rho}}, \wk | \wt) =  \E_{\Wk}\biggl[&\max_{a \in \mathcal{A}_{s_{\rho}}} f({s_{\rho}}, a |\wt  \Wk)\biggr] \\
& - \max_{a \in \mathcal{A}_{s_{\rho}}} f(s_{\rho}, a | \wt) \: ,  \\ 
\end{align*}
where $\wk$ specifies the state-actions in $\Wk$, that is, $\Wk = \{(\wbar_i, O_{\wbar_i i})\}_{i=1}^k$ where $\wbar_i$ is the $i$th element of $\wk$.
\label{voc}
\end{definition}

We refer to computation-policies that choose computations based on greedy
maximization one by one (i.e., $k=1$) of $\VOC$ as $\VOC(\phi_n)$-greedy and
$\VOC(\psi_n)$-greedy depending on which value function is utilized. We assume
these policies stop if and only if $\forall \wbar: \VOC_{f}({s_{\rho}}, \wbar |
\wt) = 0$. Our greedy policies consider and select computations one-by-one.
Alternatively, one can perform a forward search over future computation
sequences, similar to the search over future actions sequences in
\citet{guez2013}. However, this adds another meta-level to our metareasoning
problem; thus, further increasing the computational burden.

We analyze these greedy policies in terms of the $\emph{Bayesian simple
regret}$, which we define as the difference between two values. The first is
the maximum value the agent could expect to reap assuming it can perform
infinitely many computations, thus resolving all the uncertainty, before
committing to an immediate action. Given our formulation, this is identical to
$\E_{Q_0^*|\wt}\left[\max_{a \in \mathcal{A}_{s_{\rho}}}
\Upsilon_n({s_{\rho}},a | \wt)\right]$ and thus is independent of the agent's
action policy and the (future) computation policy for a given knowledge state
$\wt$. Furthermore, it remains constant in expectation as the knowledge state
expands. The second term in the regret is the maximum static/dynamic action
value assuming the agent cannot expand its knowledge state before taking an
action.

\begin{definition}
Given a knowledge state $\wt$, we define \textbf{Bayesian simple regret} at state ${s_{\rho}}$ as
\begin{equation*} 
R_f({s_{\rho}}, \wt)=  \E\left[\max_{a} \Upsilon_n({s_{\rho}},a | \wt)\right]- \max_{a} f({s_{\rho}},a | \wt) \: ,
\end{equation*}
where $f \in \{\phi_n,\psi_n\}$. 
\end{definition}

Based on this definition, we have the following.

\begin{proposition} $\VOC(\phi_n)$-greedy and $\VOC(\psi_n)$-greedy choose the computation that maximize expected decrease in $R_{\phi_n}({s_{\rho}}, \wt)$ and $R_{\psi_n}({s_{\rho}}, \wt)$ respectively.
\label{prop:voc_regret}
\end{proposition}
The proof is provided in the appendix.

We refer to policies that choose the regret-minimizing computation as being
\emph{one-step optimal}. Note that this result is different than what is
typically referred to as myopic optimality. Myopia refers to considering
the impact of a single computation only, whereas $\VOC(\psi)$-greedy policy
accounts for the impact of possible future computations that succeed the
immediate action.

\begin{proposition} Given an infinite computation budget, $\VOC(\phi_n)$-greedy and $\VOC(\psi_n)$-greedy policies will find the optimal action at the root
state.
\end{proposition}
\begin{hproof}
Both policies will perform all computations infinitely many times as shown for
the flat setting in \citet{ryzhov2012}. Thus, dynamic and static values at the
leaves (ie, for $\Gamma_n({s_{\rho}})$) will converge to the true optimal
values (given by $Q^*$), so will the downstream values.
\end{hproof}
We refer to such policies as being \emph{asymptotically optimal}.

\subsection{ALTERNATIVE VOC DEFINITIONS}
A common \citep{wefald1991,keramati2011,hay2012,tolpin2012} formulation for
the value of computation is
\begin{align}
\VOCp_{f}(s_{\rho}, \wk | \wt) =  &\E_{\Wk}\biggl[\max_{a \in \mathcal{A}_{s_{\rho}}} f(s_{\rho}, a |\wt  \Wk) \nonumber \\
& - f(s_{\rho}, \alpha | \wt  \Wk) \biggr] \: ,
\label{other_voc}
\end{align}
where $\alpha \coloneqq \arg \max_a f(s_{\rho}, a | \wt)$ and $f$ is a value
function as before.

The difference between this and
Definition~\ref{voc} is that the second term in $\VOCp$ 
conditions $f$ also on $\Wk$. This might seems intuitively
correct. $\VOCp$ is positive if and only if the policy at $s_{\rho}$
changes with some probability, that is, $P(\arg \max_{a \in
\mathcal{A}_{s_{\rho}}} f(s_{\rho}, a | \wt  \Wk)
\neq \alpha) > 0$. However, this approach can be too myopic as it often takes
multiple computations for the policy to change \cite{hay2012}. Note
that, this is particularly troublesome for static values ($f = \phi_n$),
which commonly arise in methods such as UCT that estimate mean returns
of rollouts.

\begin{proposition}$\VOCp(\phi_n)$-greedy is neither one-step optimal nor asymptotically optimal.
\label{prop:vocp_suboptimal}
\end{proposition}

By contrast, dynamic value functions escape this problem.
\begin{proposition} For any $\wk$ and $\wt$ we have $\VOCp_{\psi_n}(s_{\rho}, \wk | \wt) = \VOC_{\psi_n}(s_{\rho}, \wk | \wt)$.
\label{prop:vocp_equiv_psi}
\end{proposition}
Both propositions are proved in the Appendix.

\section{VALUE OF COMPUTATION IN MCTS}

We now introduce a MCTS method based on VOC-greedy policies we introduced. For
this, as done in other information/computation-value-based MCTS methods
\cite{hay2012,tolpin2012}, we utilize UCT as a ``base'' policy---meaning we
call UCT as a subroutine to draw samples from leaf nodes. Because UCT is
adaptive, these samples will be drawn from a non-stationary stochastic process
in practice; yet, we will treat them as being i.i.d.\ .

We introduce the model informally, and provide the exact formulas and a
pseudocode in the Appendix. We assume no discounting, i.e., $\gamma =1$, and
zero immediate rewards within $n$ steps of the root node for simplicity here,
though as we show in the Appendix, the results trivially generalize.

We assume $Q_0^* \sim \mathcal{N}(\mu_0, \Sigma_0)$ where $\mu_0$ is a prior
mean vector and $\Sigma_0$ is a prior covariance matrix. We assume both these
quantities are known---but it is possible to also assume a Wishart prior over
$\Sigma_0$ or to employ optimization methods from the Gaussian process
literature (e.g., maximizing the likelihood function via gradient descent). We
assume computations return evaluations of $Q_0^*$ with added Normal noise with
known parameters. Then, the posterior value function $Q_0^* | \wt$ can be
computed in $\mathcal{O}(t)$ for an isotropic prior covariance, in
$\mathcal{O}(tm^2)$ using recursive update rules for multivariate Normal priors
, where $m = |Q_0^*|$ is the number of leaf nodes, or in $\mathcal{O}(t^3)$
using Gaussian process priors. We omit the exact form of the posterior
distribution here as it is a standard result.

For computing the $\VOC(\phi_n)$-greedy policy we need to evaluate how the
expected values at the leaves change with a candidate computation $\wbar = (s,
a)$, i.e., $\E_{\Omega}[\E_{Q_0^* |\wt \Omega}[Q_0^* | \wt \Omega]]$ where
$\Omega = (\wbar, O_{\wbar t+1})$. Note that, $O_{\wbar t+1}$ conditioned on
$\wt$, gives the posterior predictive distribution for rollout returns from
$(s,a)$, and is normally distributed. Thus, $\E_{Q_0^* | \wt \Omega}[Q_0^* |
\wt \Omega]$ is a multivariate random Normal variable of dimension $m$. The
maximum of this variable, i.e. $\max \E_{Q_0^* | \wt \Omega}[Q_0^* | \wt
\Omega]$, is a piecewise linear function in $O_{\wbar t+1}$ and thus its
expectation can be computed exactly in $\mathcal{O}(m^2 \log m)$ as shown in \citet{frazier2009b}. If an
isotropic prior covariance is assumed, the computations simplify greatly as
$\VOC_{\phi_n}$ reduces to the expectation of a truncated univariate normal
distribution, can be computed in $\mathcal{O}(1)$ given the posterior
distributions and the leaf node with the highest expected value. If the
transitions are stochastic, then the same method can be utilized whether the
covariance is isotropic or anisotropic, with an extra averaging step over
transition probabilities at each node, increasing the computational costs.

Computing the $\VOC(\psi_n)$-policy is much harder on the other hand, even for
a deterministic state-transition function, because we need to calculate the
expected maximum of possibly correlated random variables, $\E[\max Q_0^* |
\wt]$. One could resort to Monte Carlo sampling. Alternatively, assuming an
isotropic prior over leaf values, we can obtain the following by adapting
a bound on expected maximum of random variables \citep{lrr76,ross2010}:
\begin{align*}
\E[\max Q_0^* | \wt] &\le \lambda_{s_{\rho} t}\\
&\coloneqq c + \sum_{(s', a') \in \Gamma_{n}(s_{\rho})} \biggl[ (\sigma_{s'a't})^2 F_{s'a't}(c) \\
& \quad \; + (\mu_{s'a't} - c) [1 - F_{s'a't}(c) ] \biggr] 
\end{align*}
where $\mu_{s'a't}$ and $\sigma_{s'a't}$ are posterior mean and variances, that
is $Q_0^*(s', a') | \wt \sim \mathcal{N}(\mu_{s'a't}, (\sigma_{s'a't})^2)$ , $F_{s'a't}$
is the CDF of $Q_0^*(s', a') | \wt$, and $c$ is a real number. The tightest bound is
realized for a $c$ that satisfies $\sum_{(s', a') \in \Gamma_{n}(s_{\rho})} [1
- F_{s'a't}(c)] = 1$, which can be found by root-finding methods.

The critical question is then how $\lambda_{s_{\rho} t}$ changes with an
additional sample from $(s', a')$. For this, we use the local
sensitivity, $\partial
\lambda_{s_{\rho} t} / \partial n_{s'a't}$ as a proxy, where $n_{s'a't}$ is the
number of samples drawn from $(s', a')$ until time $t$. We give the
closed form equation for this partial derivative along with some of its
additional properties in the Appendix.
Then we can approximately compute the $\VOC(\psi_n)$-greedy policy by
choosing the computation that maximizes the magnitude of $\partial
\lambda_{s_{\rho} t} / \partial n_{s'a't}$. This approach only works if
state-transitions are deterministic as it enables us to collapse the
root action values into a single $\max$ of leaf values. If the state
transitions are stochastic, this is no longer possible as averaging over
state transitions probabilities is required. Alternatively, one can
sample deterministic transition functions and average $\partial
\lambda_{s_{\rho} t} /
\partial n_{s'a't}$ over the samples as an approximation.

Our $\VOC$-greedy MCTS methods address important limitations of
$\VOCp$-based methods \citep{tolpin2012,hay2012}. $\VOC$-greedy does not
suffer from the early stopping problem that afflicts $\VOCp$-based. It
is also less myopic in the sense that it can incorporate the impact of
computations that may be performed in the future if dynamic value
functions are utilized. Lastly, our proposal extends $\VOC$ calculations
to non-root actions, as determined by $n$.

\section{EXPERIMENTS}

We compare the $\VOC$-greedy policies against UCT \citep{kocsis2006}, VOI-based
\citep{hay2012}, Bayes UCT \citep{tesauro2010}, and Thompson sampling for MCTS
(DNG-MCTS) \citep{bai2013} in two different environments, bandit-trees and peg solitaire, where the environment dynamics are provided to each method.

Bayes UCT computes approximate posterior action values and uses a rule similar
to UCT to select child nodes. DNG-MCTS also estimates the posterior action
values but instead utilizes Thompson sampling recursively down the tree. We use
the same conjugate Normal prior structure for the Bayesian algorithms:
$\VOC$-greedy, Bayes UCT, and DNG-MCTS\footnote{In the original paper
\citep{bai2013}, the authors use Dirichlet-Normal-Gamma priors, but we resort
to Normal priors to preserve consistency among all the Bayesian policies.}. The
prior and the noise parameters are tuned for each method via grid search using
the same number of evaluations, as well as the exploration parameters of UCT
and VOI-based.

VOI-based, $\VOC$-greedy, and Bayes UCT are hybrid methods, using one set of
rules for the top of the search tree and UCT for the rest. We refer to this top
part of the tree as the partial search tree (PST). By construction, VOI-based
utilizes a PST of height $1$. We implement the latter two methods using PSTs of
height $4$ in bandit-trees and of $2$ in peg solitaire. These heights are determined
based on the branching factors of the environments and the total computation budgets,
such that each leaf node is sampled a few (5-8) times on average.
For the experiments we
explain next, we tune the hyperparameters of all the policies using grid search.

\subsection{BANDIT-TREES}

\begin{figure}[h]
\centering
\includegraphics[width=0.32\textwidth]{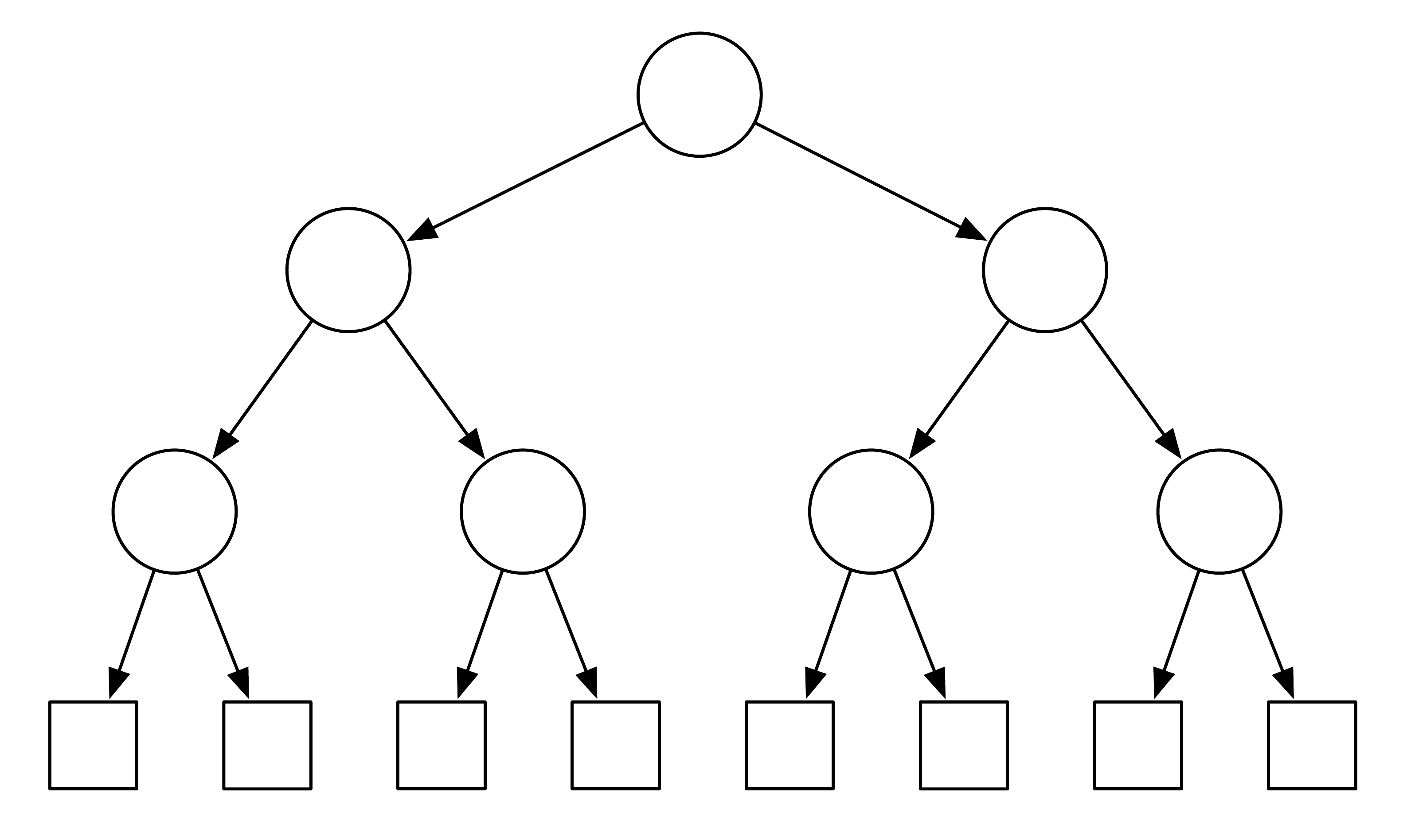}
\caption{A bandit tree of height $3$. Circles denote states, and squares denote bandit arms.}
\label{bandit-tree}
\end{figure}

The first environment in which we evaluate the MCTS policies is an MDP composed
of a complete binary tree of height $d$, similar to the setting presented in
\citet{tolpin2012} but with a deeper tree structure and stochastic transitions.
The leaves of the tree are noisy ``bandit arms'' with unknown distributions.
Agents perform ``computations'' to draw samples from the arms, which is
analogous to performing rollouts for evaluating leaf values in MCTS. At 
each state, the agents select an action from $\A = \{\texttt{LEFT},
\texttt{RIGHT}\}$ (denoting the desired subtree of height $d-1$) and
transition there with probability $.75$ and to the other subtree with
probability $.25$. In Figure~\ref{bandit-tree}, we illustrate a bandit
tree of height $3$.

At each time step $t$, agents sample one of the arms, and update their value
estimates at the root state $s_{\rho}$. We measure the simple
objective\footnote{We call it objective regret because it is based on the
ground truth ($Q^*$) as opposed to Bayesian regret, which is based on the
estimates of the ground-truth.} regret at state $s_{\rho}$ at $t$, which we
define as $\max_{a \in \A} Q^*(s_{\rho}, a) - Q^*(s_{\rho}, \pi_t(s_{\rho}))$,
for a deterministic policy $\pi_t: s_{\rho} \rightarrow \A$ which depends on
the knowledge state acquired by performing $t$ many computations.

We sample the rewards of the bandit arms from a multivariate Normal
distribution, where the covariance is obtained either from a radial basis
function or from a white noise kernel. The noise of arms/computations follow an
i.i.d. Normal distribution. We provide the exact environment parameters in the
Appendix.

Figure~\ref{fig:bt}.a shows the results in the case with correlated bandit
arms. These correlations are exploited in our implementation of
$\VOC(\phi_n)$-greedy (via an anisotropic Normal prior over the leaf values of
the PST). Note that, we aren't able to incorporate this extra assumption in
other Bayesian methods. Bayes UCT utilizes a specific approximation for
propagating values up the tree. Thompson sampling would require a prior over
all state-actions in the environment which is complicated due to the
parent-child dependency among the nodes as well as computationally prohibitive.
Because computing the $\VOC(\psi_n)$-greedy policy is very expensive if state
transitions are stochastic, we only implement $\VOC(\phi_n)$-greedy for this
environment, but implement both for the next environment.

We see that $\VOC(\phi_n)$-greedy outperforms all other methods. Note that this
is a low-sample density setting: there are $2^7=128$ bandit arms and each arm
gets sampled on average once as the maximum budget (see x-axis) is $128$ as
well. This is why many of the policies do not seem to have converged to the
optimal solution. The outstanding performance of $\VOC(\phi_n)$-greedy is
partially due to its ability of exploiting correlations. In order to control
for this, Figure~\ref{fig:bt}b shows the results in a case in which the bandit
rewards are actually uncorrelated (i.e., sampled from an isotropic Normal
distribution). As we can see $\VOC(\phi_n)$-greedy and Bayes UCT performs equally well, and better than
the other policies. This implies that the good performance of
$\VOC(\phi_n)$-greedy does not depend wholly on its ability to exploit the
correlational structure.

\begin{figure}[t]
\centering
\subfigure[Bandit-trees with correlated bandit arms.]{
\includegraphics[width=0.49\textwidth]{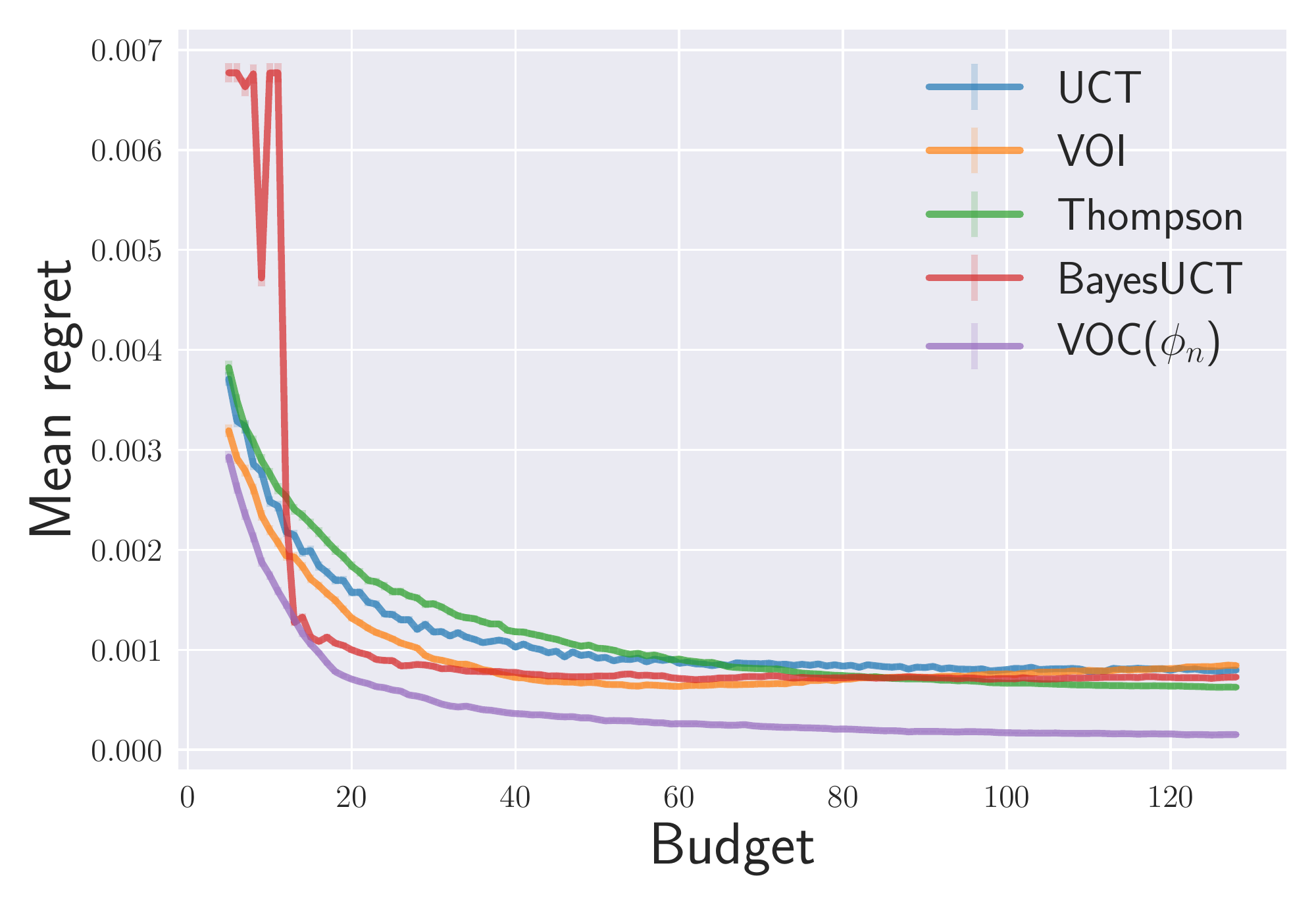}
}
\hfill
\subfigure[Bandit-trees with uncorrelated bandit arms.]{
\includegraphics[width=0.49\textwidth]{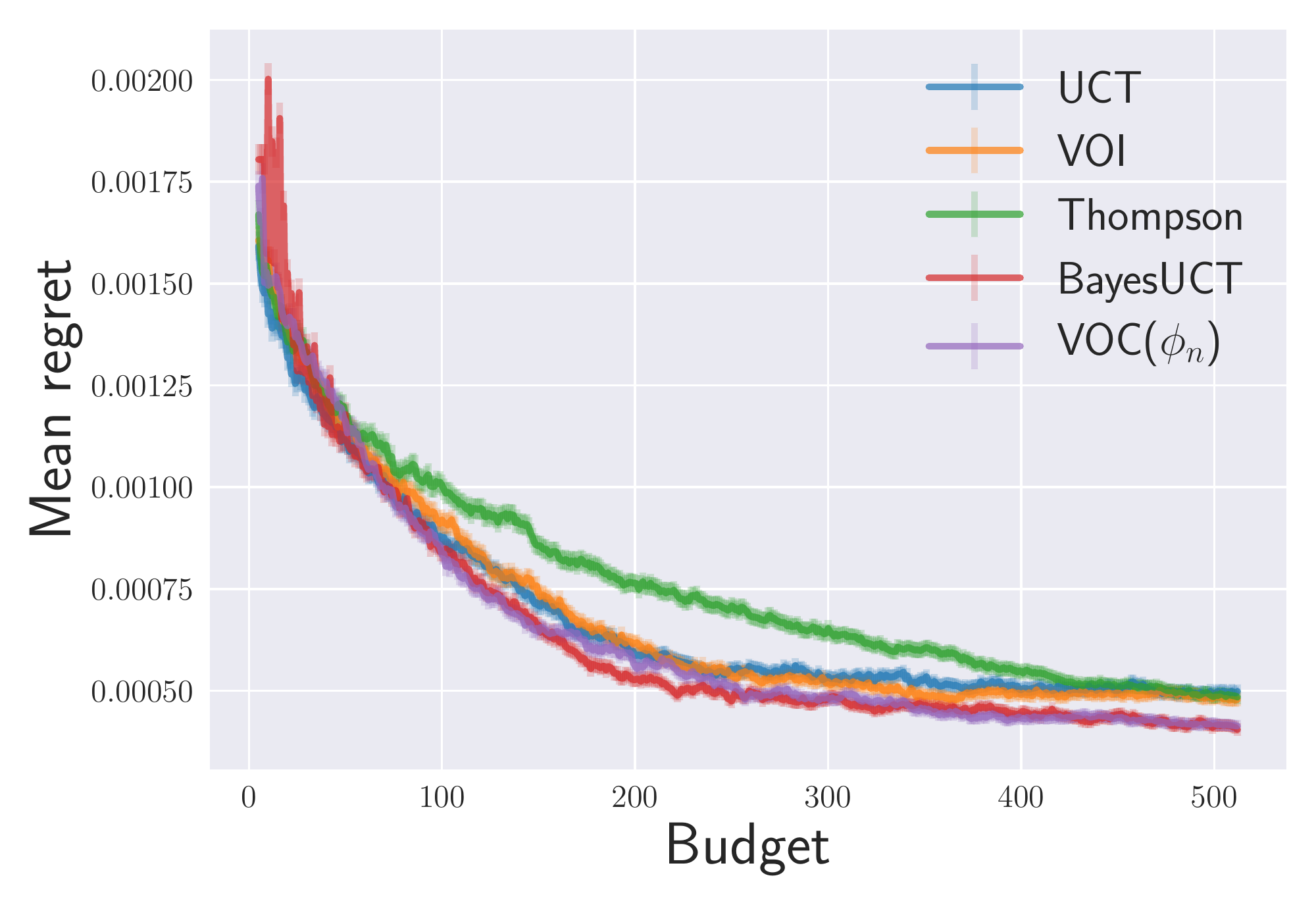}
}
\hfill
\caption{Mean regret as a function of the computation budget for bandit-trees with \emph{correlated} (panel a) \emph{uncorrelated} (panel b) expected bandit rewards, averaged over $10k$ and $5k$ random bandit reward seeds respectively.}
\label{fig:bt}
\vspace{-2em}
\end{figure}
 
\subsection{PEG SOLITAIRE}

Peg solitaire---also known as Solitaire, Solo, or Solo Noble---is a
single-player board game, where the objective for our purposes is to remove as
many pegs as possible from the board by making valid moves.
We use a $4 \times 4$ board, with $9$ pegs randomly placed. 

\begin{figure}[h]
\centering
\includegraphics[width=0.49\textwidth]{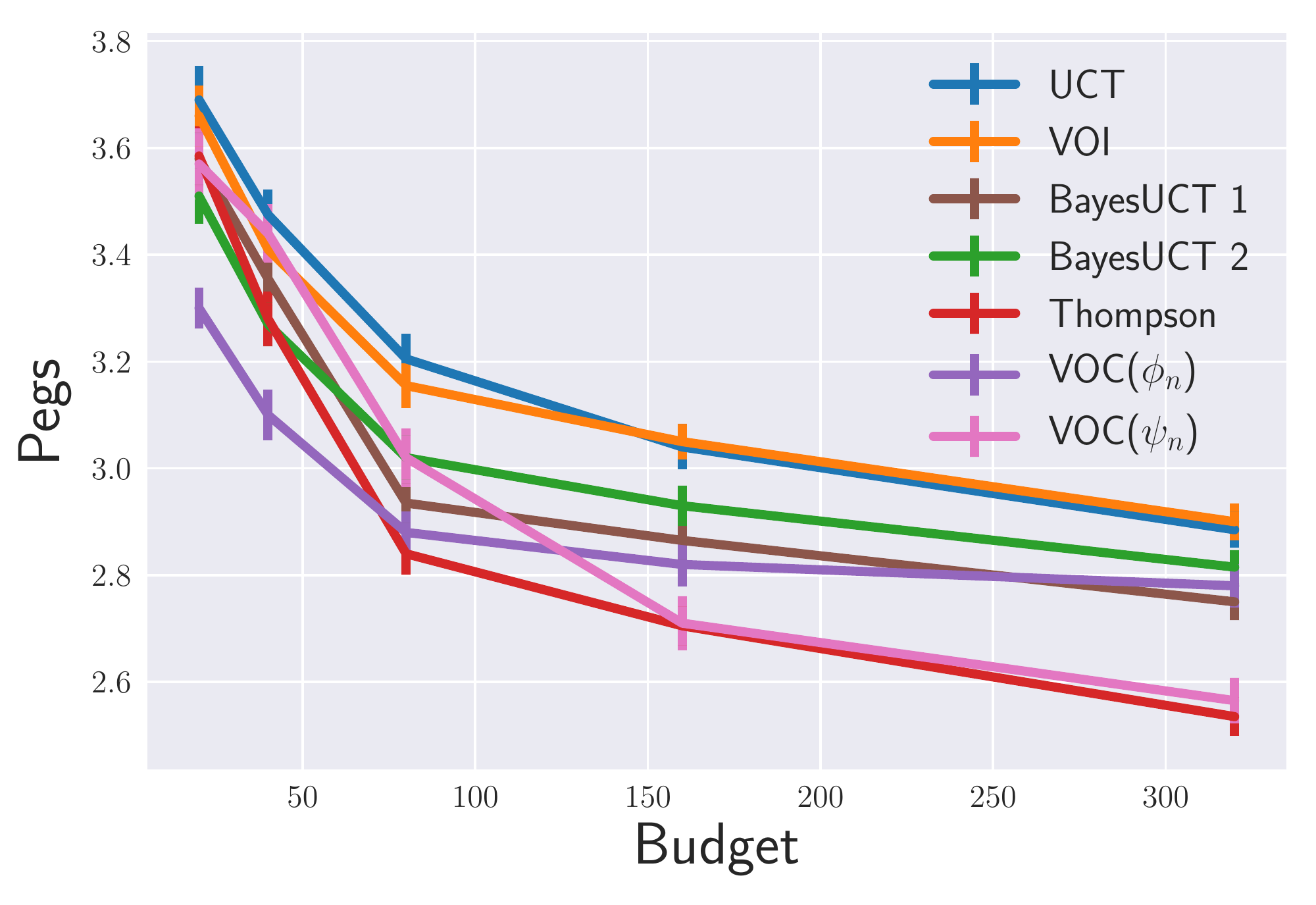}
\caption{The average number of pegs remaining on the board as a function of the computation budget, averaged over $200$ random seeds. The bars denote the mean squared errors.}
\label{ps}
\end{figure}

In the implementation of $\VOC$-greedy policies, we assume an anisotropic prior
over the leaf nodes. As shown in Figure~\ref{ps}, $\VOC(\phi_n)$-greedy has the
best performance for small budget ranges, which is in line with our intuition
as $\phi_n$ is a more accurate valuation of action values for small computation
budgets. For large budgets, we see that $\VOC(\psi_n)$-greedy performs as well
as Thompson sampling, and better than the rest.

\section{DISCUSSION}

This paper offers principled ways of assigning values to actions and
computations in MCTS. We address important limitations of existing methods by
extending computation values to non-immediate actions while accounting for the
impact of non-immediate future computations. We show that MCTS methods that
greedily maximize computation values have desirable properties and are more
sample-efficient in practice than many popular existing methods. The major
drawback of our proposal is that computing $\VOC$-greedy policies might be
expensive, and may only worth doing so if rollouts (i.e., environment
simulations) are computationally expensive. 
That said, we believe efficient derivates of our methods are possible, for instance by using graph neural networks to directly learn static/dynamic action or computation values. 

Practical applications aside, the study of computation values
might provide tools for a better understanding of MCTS policies, for instance,
by providing notions of regret and optimality for computations, similar to what
already exists for actions (i.e., $Q^*$).

\paragraph{Acknowledgements.} We are grateful to Marcus Hutter, whose feedback significantly improved our presentation of this work. We also thank the anonymous reviewers for their valuable feedback and suggestions.
ES conducted this work during a research visit to the Gatsby Unit at University College London and during his studies at the Bernstein Center for Computational Neuroscience Berlin. 
PD was funded by the Gatsby Charitable Foundation, the Max Planck
Society, and the Alexander von Humboldt Foundation.


\bibliographystyle{plainnatnourl}

\bibliography{es_pd.bib} 

\appendix
\twocolumn[
]

\section{VALUE OF COMPUTATION IN MCTS}

If the state transition function is deterministic, then static and dynamic
value computations simplify greatly:
\begin{align}
\psi_n(s,a | \wt) &= \E\left[\max_{(s', a') \in \Gamma_n(s)} Z(s', a' | \wt)\right] \label{psi_simple} \\
\phi_n(s,a | \wt) &= \max_{(s', a') \in \Gamma_n(s)}\E\left[ Z(s', a' | \wt)\right] \label{phi_simple} \: ,
\end{align}
where $Z(s',a' | \wt) \coloneqq r_1 + \gamma r_2 + \gamma^2 r_3 + \dots +
\gamma^n (Q_0^*(s', a') | \wt)$, is the posterior leaf values scaled by $\gamma^n$ and shifted by the discounted immediate rewards ($r_i$) along the path from 
$s$ to $s'$.

In this `flat' case, $\VOC(\phi_n)$-greedy policy is equivalent to a knowledge
gradient policy, details of which can be found in
\citet{frazier2009b,ryzhov2012} for either isotropic or anisotropic Normal
$Q_0^*$. On the other hand, $\VOC(\psi_n)$-greedy policy has not been studied
to the best of our knowledge. Computing the expected maximum of random
variables is generally hard, which is required for $\psi_n$. Below, we offer a
novel approximation to remedy this problem.

\subsection{COMPUTING $\VOC(\psi_n)$}
We utilize a bound \citep{lrr76} that enables us to get a
handle on $\psi_n$. This asserts,
\begin{equation*}
\psi_n(s,a | \wt) \le  c + \sum_\nsr \int_c^\infty \left[1 - F_{s'a't}(x) \right] dx \label{eq:lrr}
\end{equation*}
for any $c \in \mathbb{R}$, where $F_{s'a't}$ is the CDF of $Z(s',a'| \wt)$.
This bound does not assume independence and holds for any correlation structure
by assuming the worst case.
Furthermore, the inequality is true for all $c$. However, the tightest bound is
obtained by differentiating the RHS with respect to $c$, and setting its
derivative to zero, which in turn yields $\sum_\nsr \left[1 - F_{s'a't}(c)
\right] = 1 $ Thus, the optimizing $c$ can be obtained via line search methods.

If $Z(\cdot, \cdot | \wt)$ is distributed according to a multivariate
(isotropic or anisotropic) Normal distribution, then we can eliminate the
integral \citep{ross2010}:
\begin{align*}
\psi_n(s,a | \wt) \le \lambda_{sat} \coloneqq c + &\sum_{(s', a') \in \Gamma_{n}(s_{\rho})} \biggl[ (\sigma_{s'a't})^2 F_{s'a't}(c) \\ 
&+ (\mu_{s'a't} - c) [1 - F_{s'a't}(c) ] \biggr] 
\end{align*}
where $\mu_{s'a't}$ and $\sigma^2_{s'a't}$ are posterior mean and variances, that
is $Z^*(s', a' | \wt) \sim \mathcal{N}(\mu_{s'a't}, (\sigma_{s'a't})^2)$. 

If we further assume an isotropic Normal prior with mean $\mu_{s'a'0}$ and scale $\sigma_{s'a'0}$, and observation noise $\epsilon_i \sim \mathcal{N}(0, \sigma^2)$ i.i.d. for $i=1,2,\dots,t$, then we get the posterior mean and scale as
\begin{align*}
\mu_{s'a't} &= \frac{n_{s'a't} \hat{o}_{s'a't} / \sigma^2 + \mu_{s'a'0}/ \sigma_{s'a't}^2 }{n_{s'a't}/ \sigma^2 + 1 / \sigma_{s'a't}^2} \: , \\ 
\sigma_{s'a't} & = n_{s'a't} / \sigma_{s'a't}^2 + 1/\sigma^2 \: ,
\end{align*}

where $\hat{o}_{s'a't}$ is the mean trajectory rewards obtained from $(s', a')$ and $n_{s'a't}$ is the number of times a sample is drawn from $(s', a')$.
Then keeping $c$ fixed, we can estimate the ``sensitivity'' of $\lambda_{sat}$
with respect to an additional sample from $(s', a')$ with
\begin{equation*}
\frac{\partial \lambda_{sat}}{\partial n_{s'a't}} = \frac{\partial \lambda_{sat} }{\partial \sigma_{s'a't}}
\frac{d \sigma_{s'a't} }{d n_{s'a't}}
+  \frac{\partial \lambda_{sat} }{\partial \mu_{s'a't}} \frac{d \mu_{s'a't}}{d n_{s'a't}}\: .
\end{equation*}
where 
\begin{align*}
\frac{\partial \lambda_{sat} }{\partial \sigma_{s'a't}} &=  \frac{1}{\sqrt{2\pi}} \exp \left( \frac{-\left( \mu_{s'a't} - c\right)^2}{2\left(\sigma_{s'a't}\right)^{2}}  \right) \\
\frac{d \sigma_{s'a't}}{d n_{s'a't}} &= - \frac{\sigma \sigma_{sa0}^{3}}{2 \left(n_{s'a't} \sigma_{sa0}^{2} + \sigma^{2}\right)^{\frac{3}{2}}}\\
\frac{\partial \lambda_{sat} }{\partial \mu_{s'a't}} &= \frac{1}{2} \left(1 +  \operatorname{erf}{\left (\frac{\sqrt{2} \left(\mu_{s'a't} -c \right)}{2 \sigma_{s'a't}} \right )} \right) \\
\frac{d \mu_{s'a't} }{d n_{s'a't}} &= \frac{\sigma^{2} \sigma_{sa0}^{2} \left(- \mu_{sa0} + \hat{o}_{s'a't}\right)}{(n_{s'a't})^{2} \sigma_{sa0}^{4} + 2 n_{s'a't} \sigma^{2} \sigma_{sa0}^{2} + \sigma^{4}}
.
\end{align*}
We can then compute and utilize $\partial \lambda_{sat} / \partial n_{s'a't}$
as a proxy for the expected change in $\lambda_{sat}$. Because $\lambda_{sat}$
is an upper bound, we find that this scheme works the best when the priors are
optimistic, that is $\mu_{sa0}$ is large. In fact, as long as the prior mean is
larger than the empirical mean, $\mu_{sa0} > \hat{o}_{s'a't}$, we have
$\partial \lambda_{sat} / \partial n_{s'a't} < 0$. Then we can safely choose
the best leaf to sample from via $\arg \min_{\nsr}\left[ \max_{a \in \As}
\lambda_{sat} \right]$. We use this scheme when implementing
$\VOC(\psi_n)$-greedy in peg solitaire and confirmed that the results are
nearly indistinguishable from calculating $\VOC(\psi_n)$-greedy by drawing
Monte Carlo samples in terms of the resulting regret curves.

\section{VOC-GREEDY ALGORITHMS}

We provide the pseudocode for VOC-greedy MCTS policy in
Algorithm~\ref{alg:voc-greedy}. Throughout our analysis of this policy, 
we assume an infinite computation budget $B$.

\begin{algorithm}[h]
\SetAlgoLined
\KwIn{Current state $s_\rho$}
\KwIn{Maximum computation budget $B$}
\KwOut{Selected action to perform}
Create a partial search graph/tree by expanding state $s$ for $n$
steps \label{alg:tree}\;
Initialize the leaf set $\Gamma_n(s_\rho)$ \;
Initialize a partial function $U$, that maps states to UCT trees \;
$t \leftarrow 0$ \; 
$\wt \leftarrow \epsilon$  \tcc*[r]{empty sequence}
 \Repeat{$\max_{\wbar} \textrm{VOC}_{\phi_n/\psi_n}(s_\rho, \wbar | \wt) < 0$ \textbf{or} $t \ge B$}{
  $(s^*, a^*) = \arg \max_{\wbar} \textrm{VOC}_{\phi_n/\psi_n}(s_\rho, \wbar | \wt)$ \label{alg:comp_pol}\;
  $s^\dagger \sim \mathcal{P}_{s^*\cdot}^{a^*}$ \;
  \If{$U(s^\dagger)$ is not defined}{
  Initialize a UCT-tree rooted at $s^\dagger$ \;
  Define $U(s^\dagger)$, which maps to the UCT-tree from the previous step \;
}
Obtain sample $o_{s^* a^* t}$ by expanding $U(s^\dagger)$ and perform a roll-out \; 
$\omega_{t+1} \leftarrow (s^*, a^*, o_{s^* a^* t})$ \;
$\omega_{1:{t+1}} \leftarrow \wt \omega_{t+1}$ \;
$t \leftarrow t + 1$ \;
}
\Return $\arg \max_{a \in \mathcal{A}_{s_\rho}} \phi_n/\psi_n(s_\rho,a | \wt)$ \label{alg:pol} \; 
\caption{$\textrm{VOC}(\phi_n/\psi_n)$-greedy for MCTS}
\label{alg:voc-greedy}
\end{algorithm}

\subsection{TIME COMPLEXITIES}

Computational complexity of $\VOC$-greedy methods depend on a variety of factors, including the prior distribution of the leaf values, stochasticity, use of static vs dynamic values. Here, we discuss the time complexity of computing the $\VOC(\phi)$-greedy policy with a conjugate Normal prior (with known variance) in MDPs with deterministic transitions.

The posterior values can be updated incrementally in constant time if the prior is isotropic Normal and in $O(m^2)$ if it is anisotropic, where $m$ is the number of leaf nodes \cite{Barber:2012}.
Given the posterior distributions, the value of a computation can be computed in $O(m)$ in the isotropic case and in $O(m^2 \log m)$ in the anisotropic case. We refer the reader to \cite{frazier2009b} for further details, as the analysis done for bandits with correlated Normal arms do apply directly.

\section{PROOFS}

\subsection{PROOF OF PROPOSITION~\ref{prop:bounds}}

Let us consider the ``base case'' of $n=1$ and define a higher-order function $g$, capturing the $1$-step Bellman optimality equation for a state-action $(s, a)$:
\[
g(h) \coloneqq \sum_{s'}\mathcal{P}_{ss'}^{a} \left[ \mathcal{R}_{ss'}^{a} +
\gamma \max_{a'} h(s', a') \right] \: .
\]
Then $\phi_1(s, a | \wt) = g(\E[Q^*_0 | \wt])$ and $\psi_1(s, a | \wt) = \E[g(Q^*_0 | \wt)]$. Because $g$ is a convex function, we have $\psi_1(s, a | \wt) \ge \phi_1(s, a | \wt)$ by Jensen's inequality. We use this to prove the upper bound in Proposition~\ref{prop:bounds}:
\begin{align*}
\psi_1(s, a | \wt) &= \E_\Wk[\psi_1(s, a | \wt\Wk)] \\
& \ge \E_\Wk[\phi_1(s, a | \wt\Wk)] \: ,
\end{align*}
where the first inequality is due to Equation~\ref{consistency}.
For the lower bound, we have 
\begin{align*}
\E_\Wk[\phi_1(s, a | \wt\Wk)] &= \E_\Wk[g(\E[Q^*_0 | \wt\Wk])] \\
&\ge g(\E_\Wk[\E[Q^*_0 | \wt \Wk)]] \\
& = g(\E[Q^*_0 | \wt]) \\
& = \phi_1(s, a | \wt)  \: .
\end{align*}
These inequalities also hold for $n > 1$ for the same reasons. We omit the proof.

\subsection{PROOF OF PROPOSITION~\ref{prop:voc_regret}}
Let $\wopt$ denote the optimal candidate computation (of length 1), which
minimizes Bayesian simple regret in expectation in one-step. That is, 
\[
\wopt \coloneqq \arg \min_\wbar \E_\Omega[R_f({s_{\rho}}, \wt  \Omega)]
\] where
$\Omega \coloneqq (\wbar, O_{\wbar t+1})$ is the closure corresponding to $\wbar$.
Then, we subtracting $R_f({s_{\rho}}, \wt)$, we get
\[\wopt = \arg \min_\wbar\left[  \E_{\Omega}[R_f({s_{\rho}}, \wt
\Omega)] - R_f({s_{\rho}}, \wt)  \right] \: .
\] 
The first terms of the regrets cancel out as
$\E_{Q_0^*|\wt}\left[\max_{a \in \mathcal{A}_{s_{\rho}}}
\Upsilon_n({s_{\rho}},a | \wt)\right] = \E_{\Omega} \E_{Q_0^*|\wt
\Omega}\left[\max_{a \in \mathcal{A}_{s_{\rho}}} \Upsilon_n({s_{\rho}},a | \wt
\Omega)\right]
$.
Thus, we end up with, 
\[ \wopt = \arg \min_\wbar \biggl[ -
\E_\Omega \left[\max_a f({s_{\rho}},a |\wt \Omega) \right] + \max_{a}
f({s_{\rho}},a | \wt) \biggr] \: ,
\]
or equivalently $\wopt = \arg \max_\wbar \VOC_f({s_{\rho}}, \Omega | \wt)$.

\subsection{PROOF OF PROPOSITION~\ref{prop:vocp_suboptimal}}

\begin{figure}[h]
\begin{center}
\includegraphics[width=0.45\textwidth]{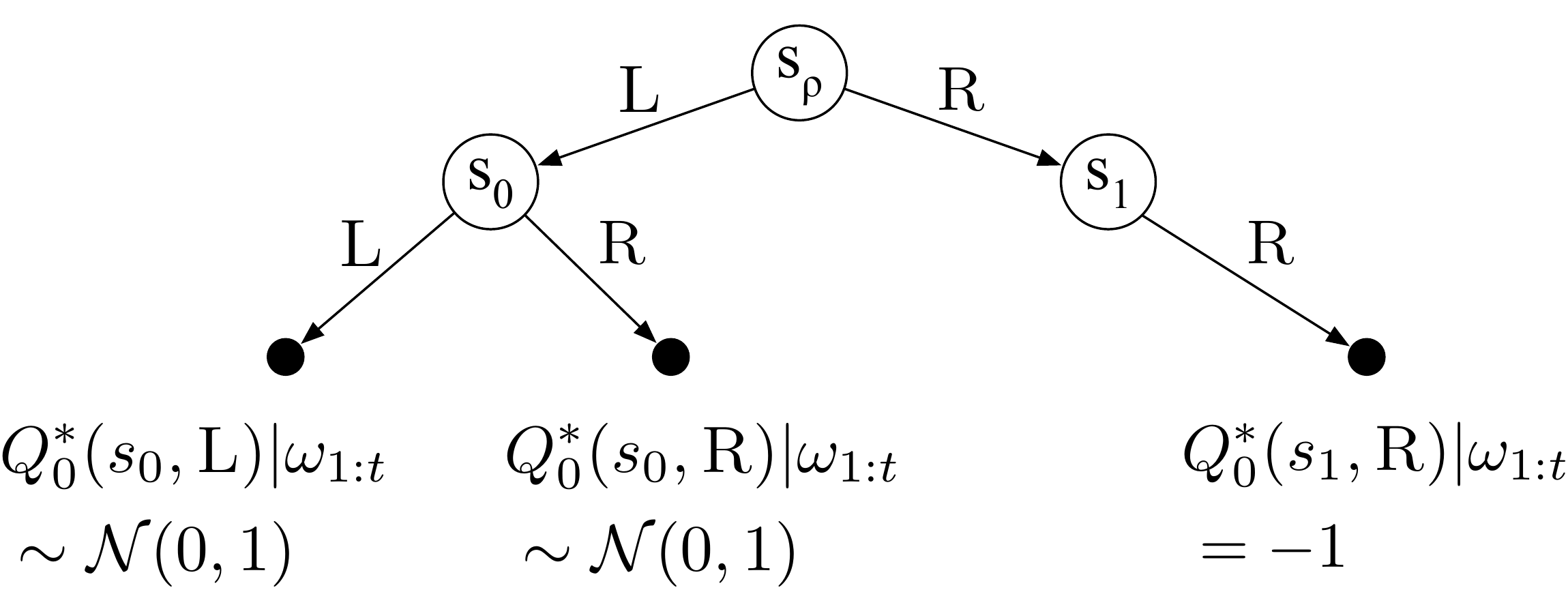}
\caption{A search graph, where $\VOCp(\phi_n)$-greedy stops early.}
\label{counter_ex}
\end{center}
\end{figure}

Consider the following $2$-step (i.e., $n=2$) search tree shown in
Figure~\ref{counter_ex}. The deterministic transitions are shown with the
arrows, each corresponding to an action in $\mathcal{A} = \{\textsc{L},
\textsc{R}\}$. The leaves are denoted with filled circles whose posterior
values are given by $Q_0^*(\cdot, \cdot) | \wt$. The root state is shown as
$s_{\rho}$ with its immediate successors as $s_0$ and $s_1$. Assume
$\gamma=1$ and all shown actions yield $0$ immediate rewards. Finally, assume
the posterior distribution of the leaf values are as in Figure~\ref{counter_ex}, and they are pairwise independent.

In this case, we can see that no single sample from the leafs can result
in a policy change at $s_{\rho}$ since we would need to sample both of
the leaves of the left subtree at least once for the policy at the root
to change from $\textsc{L}$ to $\textsc{R}$. Therefore, $\VOCp_{\phi_2}$
is zero for all possible computations here, and thus stops early, not
achieving neither one-step nor asymptotic optimality. In contrast,
$\VOC_{\phi_2}$ is greater than zero for computations concerning the
left subtree.

\subsection{PROOF OF PROPOSITION~\ref{prop:vocp_equiv_psi}}

We need to show the equality of the second term in Equation~\ref{other_voc}
to the second term of $\VOC$ as we defined in Definition~\ref{voc}. First
observe that $\E_{\Wk}[\psi_n(s_{\rho}, \alpha | \wt  \Wk)] =
\psi_n(s_{\rho}, \alpha | \wt)$. Then, we can take the $\alpha$ out as
$\psi_n(s_{\rho}, \alpha | \wt) = \max_{a \in \A_{s_\rho}}\psi_n(s_{\rho}, a |
\wt)$, which is identical to the second term of our $\VOC$ definition in
Equation~\ref{other_voc}.

\section{BANDIT TREE DETAILS}
We utilizes trees of depth 7, where the agent transitions to the desired sub-tree with probability $.75$. In the correlated bandit arms case, the expected rewards of the arms are sampled from $\mathcal{N}(1/2, \Sigma)$ i.i.d. at each trial, where $\Sigma$ is the covariance matrix given by an RBF kernel with scale parameter of $1$ and the observation noise is sampled from $\mathcal{N}(0, 0.1)$ i.i.d. at each time step. In the uncorrelated case, the expected rewards are sampled from $\mathcal{U}(0.45, 0.55)$ and the observation noise is from $\mathcal{N}(0, 0.01)$. The former setting is designed to be noisier to compensate for the extra information provided by the correlations.

\end{document}